\documentclass[11pt]{article}

\usepackage[final]{acl}

\usepackage{times}
\usepackage{latexsym}
\usepackage{enumitem}

\usepackage[T1]{fontenc}

\usepackage[utf8]{inputenc}

\usepackage{microtype}

\usepackage{inconsolata}

\usepackage{graphicx}

\usepackage{amsmath}
\usepackage{amssymb}
\usepackage{booktabs}
\usepackage{multirow}
\usepackage{xspace}
\usepackage{amsthm}
\newtheorem{proposition}{Proposition}
\usepackage{xcolor}
\definecolor{darkcyan}{RGB}{0,100,100}

\newcommand{\titlename}{TAG-DLM\xspace}
\newcommand{\fulltitle}{Diffusion Language Models for \\Text-Attributed Graph Learning\xspace}

\newcommand{\methodname}{TAG-DLM}
\newcommand{\method}{\textbf{\texttt{\methodname}}\xspace}
\newcommand{\fullmethod}{\textbf{T}ext-\textbf{A}ttributed \textbf{G}raph \textbf{D}iffusion \textbf{L}anguage \textbf{M}odel\xspace}

\newcommand{\TAG}{TAG\xspace}
\newcommand{\GAT}{GAT\xspace}
\newcommand{\GNN}{GNN\xspace}
\newcommand{\GNNs}{GNNs\xspace}
\newcommand{\MPNN}{MPNN\xspace}
\newcommand{\MPNNs}{MPNNs\xspace}
\newcommand{\MDLM}{MDLM\xspace}
\newcommand{\MDLMs}{MDLMs\xspace}

\title{\titlename: \fulltitle}

\author{Lingjie Chen\thanks{Equal contribution.} \\
  UIUC \\
  \texttt{lingjie7@illinois.edu} \\\And
  Yuanchen Bei$^*$ \\
  UIUC \\
  \texttt{bei4@illinois.edu} \\\And
  Haobo Xu$^*$ \\
  UIUC \\
  \texttt{haoboxu@illinois.edu} \\\AND
  Yanjun Zhao \\
  UIUC \\
  \texttt{yanjunzh@illinois.edu} \\\And
  Yuzhong Chen \\
  VISA \\
  \texttt{yuzchen@visa.edu} \\\And
  Hanghang Tong \\
  UIUC \\
  \texttt{htong@illinois.edu} \\}

\begin{document}
\maketitle

\begin{abstract}
Text-attributed graphs (TAGs),  where each node carries a natural language description, require models to jointly reason over text and graph topology. Existing approaches often handle the two modalities separately: graph neural networks operate on shallow text features, while hybrids of LLMs and graphs use the language model mainly as a text encoder and delegate structure learning to a separate graph module. We propose \method, which unifies textual reasoning and graph message passing within a masked diffusion language model, a language model with bidirectional attention and generative decoding. For each graph instance, \method linearises a sampled local neighbourhood into a token sequence and injects graph structure through a  \textit{topology attention mask}, which realises message passing over the graph. Because the diffusion language model can both interpret and generate text, \method adapts to different tasks simply by changing the prompt, supporting node classification, link prediction, and cross-dataset transfer with no target-specific fine-tuning.
Experiments show that \method outperforms graph neural networks, graph transformers, and LLM-based baselines on all three TAG benchmarks across two tasks, improving over the strongest baseline by up to 3.9 points.
\end{abstract}

\section{Introduction}
\label{sec:intro}

Graphs whose nodes carry natural language descriptions arise across a wide range of real-world systems, such as citation networks in which each paper is represented by its title and abstract, biomedical literature graphs whose nodes describe research topics or articles, and knowledge bases in which entities are described by free-form passages~\cite{yan2023comprehensive,wang2025can}. A graph of this kind is a \textit{text-attributed graph} (\TAG): every node is annotated with a piece of text, and the edges represent relations between nodes. \TAG learning asks a model to make predictions over such graphs from both the node text and the graph topology~\cite{wang2024bridging}. We study two standard tasks: in node classification (NC), the model assigns a label to a target node; in link prediction (LP), it decides whether an edge exists between a candidate pair. In both cases the answer is specified by the prompt and must be inferred from two complementary sources of evidence, the textual content of the local neighbourhood and its connectivity. Solving these tasks well therefore requires a model that can simultaneously understand natural language and reason over graph structure. 


\begin{figure}[t]
  \centering
  \includegraphics[width=\linewidth]{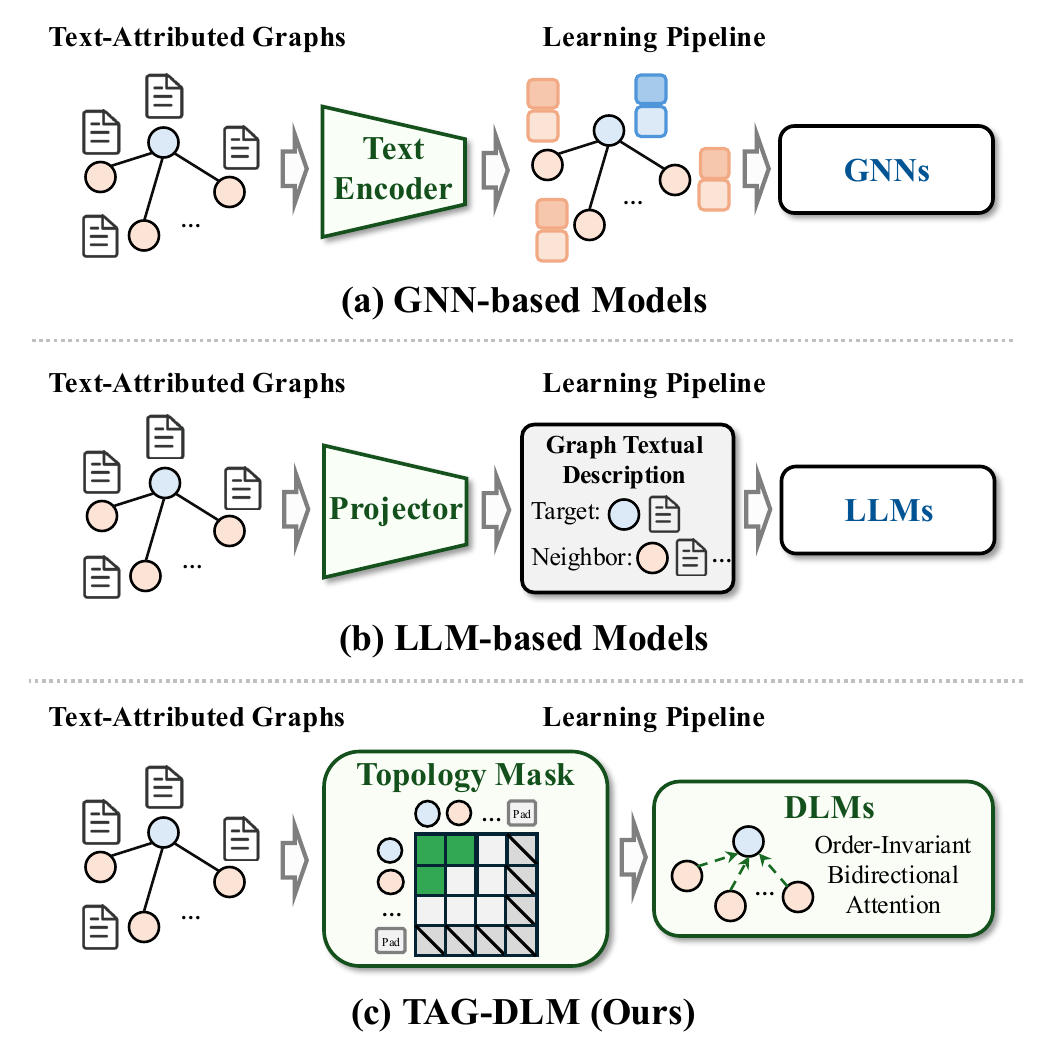}
  \vspace{-0.5em}
  \caption{Comparison between \method with existing text-attributed graph learning pipelines.}
  \vspace{-0.5em}
  \label{fig:pipeline-comparison}
\end{figure}

Existing approaches usually place language understanding and graph reasoning in separate components, which limits how the two sources of evidence interact. Graph Neural Networks (\GNNs), such as GCN~\citep{kipf2017gcn} and \GAT~\citep{velickovic2018gat} model graph structure effectively but represent node text as shallow bag of words or truncated embeddings, discarding most of the semantic content. More recent large language model (LLM) hybrids~\citep{chen2024llaga,tang2024graphgpt} treat the language model as a frozen text encoder: node texts are embedded independently, and a separate \GNN is then attached to propagate the resulting embeddings over the graph. 
This pipeline keeps the two modalities structurally separate. The language model never conditions on graph topology, and the graph module operates on these precomputed text embeddings rather than the raw node text. As a result, the contextual understanding of the LLM cannot interact with graph-level reasoning: each side sees only a degraded view of the other. A further limitation is task specificity: because the task is read out by a fixed output head, these methods must be retrained whenever the task changes.

First, \MDLMs use fully bidirectional self-attention, so information flows in all directions within a single forward pass. This suits graph prompts, where the linearised order of neighbour nodes is arbitrary and the target and its neighbours should exchange evidence regardless of that order, rather than along a fixed left-to-right direction.
Second, \MDLMs are generative: they fill in masked spans in a sequence, so the answer space can be specified by the prompt rather than by an output head specific to a task. 
These two properties motivate the key insight behind our work: \emph{graph neighbourhood relationships can be encoded directly as an attention mask}. By restricting which tokens may attend to which, we embed graph structure into the denoising process without a separate graph module or any change to the backbone. And because the answer is simply a filled-in prompt position, the same formulation expresses both NC and LP through prompt variation, with no task-specific graph head.

Building on this, we introduce \method (\fullmethod), a framework that casts \TAG learning as a masked infill problem. For a target graph instance, such as a node to classify or a candidate edge to score, \method linearises the sampled local context into a single token sequence, fills a masked answer position with an option defined by the prompt, and attaches a binary \textit{topology attention mask}. The mask enforces structured attention over the local graph context, so a single forward pass through the \MDLM backbone simultaneously reads textual evidence and propagates information along the graph, with no separate \GNN component required.

Our contributions are summarized as follows:
\begin{itemize}[leftmargin=*]
  \item \textbf{Topology attention mask as graph operator.} Graph structure is injected
    into an \MDLM purely through a binary attention mask for each sample,
    eliminating the need for a separate \GNN encoder or any architectural
    modification to the backbone.
  \item \textbf{\MDLMs for \TAG learning.}
    \method adapts the masked diffusion paradigm to graph-structured data,
    combining bidirectional language understanding with graph topology in a
    single model.
  \item \textbf{Theoretical grounding.} We prove that a single denoising
    forward pass with the topology attention mask performs attention-based message
    passing on the graph induced by the mask.
  \item \textbf{Unified graph learning evaluation.} Since \method is
    generative, node classification, link prediction, and cross-dataset
    transfer can be expressed through answer options defined by the prompt rather
    than output heads specific to a task.
\end{itemize}

\section{Preliminaries}
\label{sec:preliminary}

\subsection{Text Attributed Graphs}
\label{sec:prelim:tag}

A \emph{text attributed graph} (\TAG) is a triple $\mathcal{G} = (\mathcal{V}, \mathcal{E}, \mathcal{X})$ in which $\mathcal{V}$ is the node set, $\mathcal{E} \subseteq \mathcal{V} \times \mathcal{V}$ is the edge set, and $\mathcal{X} = \{x_v\}_{v\in \mathcal{V}}$ assigns to every node $v$ a piece of natural language text $x_v$. We consider graph prediction tasks defined by prompts over such graphs. In \emph{node classification} (NC), each node carries a discrete label $y_v \in \mathcal{Y}$ from a finite, language expressible label set $\mathcal{Y}$ (e.g.\ ``Neural Networks'', ``cs.CV''), the labels of a subset $\mathcal{V}_\mathrm{train} \subseteq \mathcal{V}$ are observed, and the goal is to predict $y_v$ for held out nodes. In \emph{link prediction} (LP), the model receives a candidate pair $(u,v)$ and predicts whether the edge belongs to the target graph under a binary answer set. Both tasks are cast as selecting an answer from a candidate set specified in the prompt. The ego graph of a target node $v$ within $k$ hops is denoted $\mathcal{N}_k(v) \subseteq \mathcal{V}$ and contains $v$ together with the nodes reachable from $v$ in at most $k$ edge steps; for pairwise tasks, we use the sampled local context around the candidate endpoints.

\subsection{Masked Diffusion Language Models}
\label{sec:prelim:mdlm}

\method uses a \emph{masked diffusion language model} (\MDLM) as its backbone~\citep{sahoo2024mdlm,nie2025llada}. Let $V$ denote the model vocabulary augmented with a special mask token \texttt{[M]} and let $x = (x_1, \ldots, x_L) \in V^L$ be a clean token sequence. The \emph{forward} process corrupts $x$ over a continuous time $t \in [0, 1]$ by independently replacing each token with \texttt{[M]} with probability $\beta(t)$, where $\beta$ is a monotone schedule with $\beta(0) = 0$ and $\beta(1) = 1$. Writing $\widetilde{x}_t$ for the sequence at time $t$, the forward kernel factorises across positions:
\begin{equation}
  q(\widetilde{x}_t \mid x)
  = \prod_{i=1}^{L}
    q\bigl(\widetilde{x}_{t,i} \mid x_i\bigr).
\end{equation}

A denoising network $f_\theta$ predicts each clean token from a corrupted sequence using \emph{bidirectional} self-attention over all positions. Training maximises a Rao Blackwellised evidence lower bound~\cite{liu2019rao} that reduces, up to a reweighting $w(t)$ that depends on time, to a masked cross entropy on the corrupted positions:
\begin{equation}
\resizebox{\linewidth}{!}{$
\mathcal{L}_{\mathrm{MDLM}}(\theta)
=
\mathbb{E}_{t,x,\widetilde{x}_t}
\left[
w(t)
\sum_{i:\widetilde{x}_{t,i}=\texttt{[M]}}
-\log f_{\theta}(x_i \mid \widetilde{x}_t, t)
\right].
$}
\label{eq:mdlm-loss}
\end{equation}
At inference time, generation proceeds by an iterative denoising loop that starts from a fully or partially masked sequence and progressively unmasks tokens by sampling from $f_\theta$. Crucially for our analysis in Section~\ref{sec:theory}, $f_\theta$ uses attention that is not causal, so the only structural restriction on information flow within a layer is whatever attention mask is supplied externally.

\subsection{Attention-Based Message Passing}
\label{sec:prelim:attention-mp}

Many graph layers can be written as message passing with learned attention over a node's neighbourhood. Given node representations $h_u^\ell$ at layer $\ell$, an attention-based message passing layer has the form
\begin{align}
  m_u^\ell
  &= \sum_{w \in \mathcal{N}(u)}
       \alpha_{uw}^\ell\, \phi_\ell(h_w^\ell), \\
  \alpha_{uw}^\ell
  &= \frac{
       \exp s_\ell(h_u^\ell, h_w^\ell)
     }{
       \sum_{r \in \mathcal{N}(u)}
       \exp s_\ell(h_u^\ell, h_r^\ell)
     }, \\
  h_u^{\ell+1}
  &= U_\ell(h_u^\ell, m_u^\ell),
\end{align}
where $s_\ell$ is an attention score, $\phi_\ell$ transforms neighbour features, and $U_\ell$ updates the node state. \GAT \citep{velickovic2018gat} is a standard instance of this family: it uses an additive attention score $s_\ell(h_u,h_w)=\operatorname{LeakyReLU}(a^\top[Wh_u\,\Vert\,Wh_w])$ and a shared linear value transform. Section~\ref{sec:theory} shows that topology masked denoising realizes this broader attention-based message passing form, with \GAT recovered as a special case.

\section{Method}
\label{sec:method}

\subsection{Overall Framework}
\begin{figure*}[t]
  \centering
  \includegraphics[width=\linewidth, trim=0 5cm 0 5cm, clip]{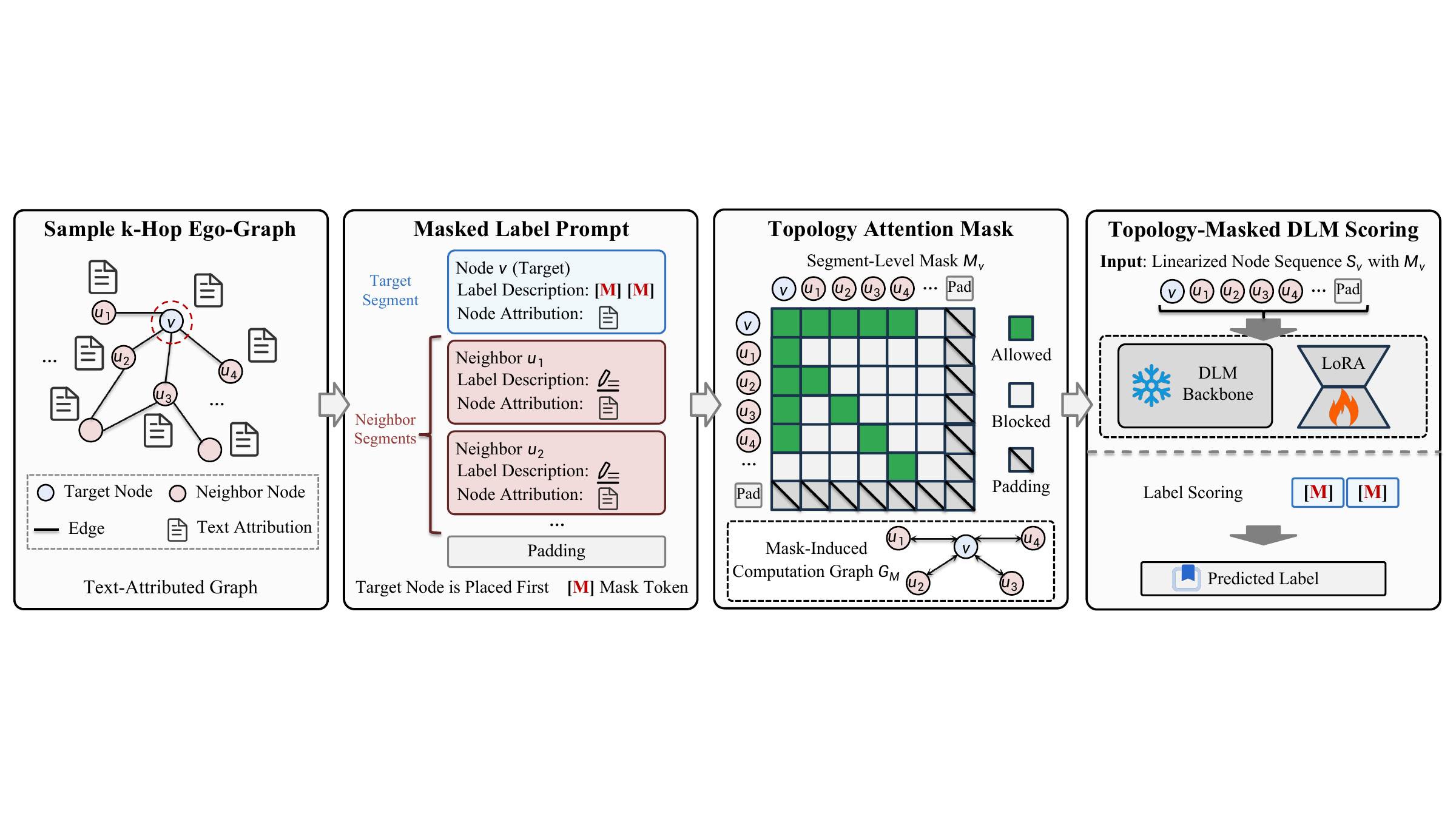}
  \vspace{-2em}
  \caption{Overview of \method. The ego graph of target node $v$ within $k$ hops is linearised into a token sequence $S_v$ with the answer position replaced by \texttt{[M]}. A binary topology attention mask $M_v$ enforces a star-shaped attention pattern: $v$ attends to all tokens; each neighbour attends only to itself and $v$. Both are fed into LLaDA-8B with LoRA tuning, which predicts the answer defined by the prompt from the answer position logits.}\vspace{-0.6em}
  \label{fig:method-overview}
\end{figure*}

\method casts graph prediction as masked label infilling over a linearised sequence that contains the target instance and its sampled local context. Graph structure is injected into self-attention through a topology attention mask rather than through a separate \GNN encoder. Whereas existing hybrids of LLMs and graphs treat the language model as a frozen text encoder attached to a graph module, \method makes the language model itself the graph operator: the only graph aware input is a binary attention mask for each sample, and the only trained parameters are LoRA adapters on the linear projections. This design gives two useful properties. First, the underlying \MDLM weights are reused without architectural changes or additional graph modules. Second, masked self-attention performs message passing on the graph induced by the mask (Section~\ref{sec:theory}), so a single bidirectional forward pass through the \MDLM reads textual evidence and propagates information along the graph. As illustrated in Figure~\ref{fig:method-overview}, \method has two core components: prompt construction from a sampled local graph context (\S\ref{sec:method:prompt}) and a topology attention mask that constrains information flow inside the \MDLM (\S\ref{sec:method:topology-mask}).

\subsection{Prompt Construction}
\label{sec:method:prompt}

\paragraph{Subgraph Linearisation.} For each target node $v \in \mathcal{V}$ in a \TAG $\mathcal{G} = (\mathcal{V}, \mathcal{E}, \mathcal{X})$, we sample an ego graph within $k$ hops, $\mathcal{N}_k(v) = \{v\} \cup \{u_1, \ldots, u_{n_v}\}$, by uniformly selecting at most $K$ neighbours per hop (default $k{=}2$, $K{=}10$). The linearised sequence is
\begin{equation}
  S_v \;=\; T(v) \;\Vert\; T(u_1) \;\Vert\; \cdots \;\Vert\; T(u_{n_v}),
\end{equation}
where $T(\cdot)$ denotes the template for each node and $\Vert$ is sequence concatenation. The target node always occupies the first segment so that its token span can be recovered directly. For pairwise prediction, we use the same construction around the candidate endpoints and encode the candidate relation in the prompt.

\paragraph{Multiple Choice Prompt.} We use the \emph{multiple choice digit} prompt format for both node classification and link prediction. The answer position is a single \texttt{[M]} token, and each candidate answer is represented by a digit token. For node classification, the digit options index the dataset label names. For link prediction, the same scoring rule is used with binary options indicating whether the candidate edge exists. Neighbour templates may include observed training labels inside square brackets, while neighbours without observed labels drop the bracket prefix. We provide a concrete prompt example in Appendix~\ref{app:prompt}.

\subsection{Topology Attention Mask}
\label{sec:method:topology-mask}

A central design choice in \method is to expose graph adjacency only through attention. The base \MDLM weights remain frozen, and graph structure enters as an input-dependent attention constraint rather than as a separate graph encoder. For the tokenised sequence $S_v$, let $|S_v|$ denote its number of tokens. We attach to every sample a binary mask $M_v \in \{0,1\}^{|S_v| \times |S_v|}$ whose rows and columns correspond to token positions in $S_v$. Conceptually, $M_v$ is a token span level graph operator: each row specifies which target or context span may read from which other span during self-attention.

The default mask follows a star-shaped attention pattern. Tokens in the target span attend to every token in $S_v$, so the target sees context from one to $k$ hops. Tokens in a neighbour span attend to themselves and to the target span, but not to other neighbours, yielding star-shaped message flow with the target as the root. Padding positions are masked in both directions, so they neither send nor receive information.
This star mask is an intentional local context abstraction for target-centred prediction. The target instance acts as the unique aggregation point and reads evidence from all sampled nodes, which keeps the same prompt and mask interface for node classification and link prediction. Suppressing direct communication among neighbours also reduces sensitivity to dataset-specific ego graph wiring, so transfer across datasets depends less on idiosyncratic local topology. Richer masks could expose more of the original ego graph structure, but we use the star mask as the default because it is simple, agnostic to the task, and stable across graph datasets.

\section{Theoretical Analysis}
\label{sec:theory}

We now show that, at any fixed diffusion time, one self-attention layer of the topology masked \MDLM denoiser implements an attention-based message passing update over the neighbourhoods induced by the topology attention mask. The argument explains why injecting the graph through attention alone, without a separate \GNN encoder, is expressive enough to subsume standard graph attention, and makes explicit which node spans exchange information. The full derivation, multihead extension, and the corollary recovering \GAT appear in Appendix~\ref{app:proof}.

\subsection{Fixed Time Denoising Operator}
\label{sec:theory:fixed-time}

The \method denoising network $f_\theta(\widetilde X_t, t, M_v)$ takes the corrupted token sequence $\widetilde X_t$ at diffusion time $t$, the topology attention mask $M_v$ for the target, and predicts the clean tokens \emph{in parallel} through stacked self-attention layers. Throughout this section, we fix $t$ and analyse a single layer, separating the temporal denoising process from graph-structured communication. The reverse diffusion process evolves the corrupted sequence across timesteps, whereas graph information flow is induced within each denoising call by the topology attention mask over token spans. Our claim is therefore local to a fixed time denoising operator: a masked self-attention layer realizes spatial message passing on the graph induced by the mask, independently of the noising and denoising dynamics over $t$.

The crucial property of the backbone is that, unlike a causal language model whose attention mask forbids tokens from attending to later positions, the \MDLM denoiser uses \emph{bidirectional} self-attention. After the topology attention mask $M_v$ is applied, the \emph{only} structural constraint on information flow within a layer is $M_v$ itself, rather than token order from left to right. This is what allows the masked denoiser to be read as graph message passing.

\subsection{Neighbourhoods Induced by the Mask}
\label{sec:theory:induced-graph}

Using the linearised sequence from Section~\ref{sec:method:prompt}, let $I_u$ denote the contiguous token span of node $u$. The topology attention mask determines which node spans can exchange information. We write $\mathcal{N}_M(u)$ for the set of nodes whose spans may be attended to by tokens in $I_u$ under $M_v$. For the star mask used in \method, the target node attends to all sampled nodes, while each node other than the target attends only to itself and the target. Thus, the message passing structure analysed below is induced by the mask, rather than by the original \TAG edges.

\subsection{Masked Denoising as Message Passing}
\label{sec:theory:proposition}

Let $h_u^\ell \in \mathbb{R}^d$ denote the node level representation obtained by pooling the token hidden states over the span $I_u$ at layer $\ell$, and let $W_Q^\ell, W_K^\ell, W_V^\ell$ be the layer-$\ell$ query, key, and value projections.

\begin{proposition}[Topology masked denoising as message passing]
\label{prop:mp}
At any fixed diffusion time $t$, one bidirectional self-attention
layer of $f_\theta$ restricted by the topology attention mask $M_v$ admits the
form
\begin{equation}
  h_u^{\ell+1}
  = U_\ell\!\left(
      h_u^\ell,\;
      \sum_{w \in \mathcal{N}_M(u)}
        \alpha_{uw}^\ell\, W_V^\ell h_w^\ell
    \right),
  \label{eq:mp-form}
\end{equation}
where the attention weight is
\begin{equation}
  \alpha_{uw}^\ell
  = \frac{
      \exp\!\bigl(s_\ell(h_u^\ell, h_w^\ell, t)\bigr)
    }{
      \sum_{r \in \mathcal{N}_M(u)}
        \exp\!\bigl(s_\ell(h_u^\ell, h_r^\ell, t)\bigr)
    },
  \label{eq:mp-alpha}
\end{equation}
$s_\ell(\cdot)$ is the attention score conditioned on the timestep, and
$U_\ell$ is the layer update formed by the residual connection, LayerNorm,
and the FFN.
Equation~\eqref{eq:mp-form} is the canonical attention-based
message passing layer over the neighbourhoods induced by the mask.
\end{proposition}

\paragraph{Proof Sketch.}
Masked self-attention can be matched term by term to an attention-based \MPNN. The topology attention mask zeros out $\alpha_{uw}^\ell$ for $w \notin \mathcal{N}_M(u)$, so the softmax in Eq.~\eqref{eq:mp-alpha} reduces to a normalisation over the induced neighbourhood. The value projection $W_V^\ell h_w^\ell$ defines the message content, the normalised weight $\alpha_{uw}^\ell$ is the message coefficient, and the masked attention sum is the aggregation operator. The residual connection together with the output projection, LayerNorm, and feed-forward block constitutes the update function $U_\ell$. Replacing token-level attention with span-pooled attention under the compatibility assumptions in Appendix~\ref{app:proof} yields Eqs.~\eqref{eq:mp-form}-\eqref{eq:mp-alpha}.\hfill\qed

\subsection{Implications}
\label{sec:theory:implications}

The proposition above has two consequences. First, choosing $s_\ell$ to be additive attention recovers the \GAT update exactly, so \GAT is a special case of \method rather than a separate design; the dot-product score used by \method generalises this to a broader family of attention \MPNNs. Second, the full reverse diffusion procedure composes timestep conditioned message passing operators over the same neighbourhoods induced by the mask, which justifies treating the iterative infill mode of \method as repeated graph reasoning rather than a separate inference mechanism.

\section{Experiments}
\label{sec:experiments}

\begin{table*}[!t]
  \centering
  \small
  \setlength{\tabcolsep}{4pt}
  \begin{tabular*}{\textwidth}{@{\extracolsep{\fill}}llcccccc@{}}
    \toprule
    & & \multicolumn{3}{c}{Node classification} & \multicolumn{3}{c}{Link prediction} \\
    \cmidrule(lr){3-5}\cmidrule(lr){6-8}
    Family & Model & Cora & PubMed & ogbn-arxiv & Cora & PubMed & ogbn-arxiv \\
    \midrule
    \multirow{6}{*}{\begin{tabular}[c]{@{}l@{}}\textit{Graph neural}\\\textit{networks}\end{tabular}}
      & GCN              & 89.48 & 93.56 & 74.25 & \underline{87.79} & 89.12 & \underline{94.67} \\
      & GraphSAGE        & 88.93 & 94.93 & 75.76 & 85.00 & 84.30 & 90.61 \\
      & GAT              & 88.93 & 92.24 & 73.80 & 86.91 & 87.59 & 91.93 \\
      & GATv2            & 89.11 & 94.68 & 76.02 & 85.59 & 86.90 & 92.02 \\
      & GIN              & 88.01 & 93.59 & 70.89 & 81.18 & 80.35 & 81.79 \\
      & MixHop           & 88.75 & 95.23 & 76.01 & 84.26 & 89.21 & 92.53 \\
    \midrule
    \multirow{4}{*}{\begin{tabular}[c]{@{}l@{}}\textit{Graph}\\\textit{transformers}\end{tabular}}
      & GraphTransformer & 88.01 & 95.03 & 76.12 & 82.79 & 86.62 & 93.43 \\
      & NodeFormer       & \underline{89.67} & 95.16 & 76.63 & 80.00 & 87.50 & 92.73 \\
      & DIFFormer        & 88.56 & 95.03 & 76.29 & 79.85 & 81.67 & 92.93 \\
      & SGFormer         & 87.64 & \underline{95.28} & 76.08 & 85.44 & 89.74 & 93.32 \\
    \midrule
    \multirow{3}{*}{\begin{tabular}[c]{@{}l@{}}\textit{LM}\\\textit{baselines}\end{tabular}}
      & RoBERTa          & 83.17 & N/A & N/A & N/A & N/A & N/A \\
      & LLaGA            & 89.22 & 95.03 & \underline{76.66} & 86.82 & \underline{91.41} & 94.15 \\
      & Frozen LLaDA-8B  & 52.65 & 51.12 & 46.99 & 52.09 & 47.89 & 48.75 \\
    \midrule
    \multicolumn{2}{l}{\method (ours)} & \textbf{90.96} & \textbf{96.30} & \textbf{76.93} & \textbf{91.62} & \textbf{95.31} & \textbf{96.55} \\
    \bottomrule
  \end{tabular*}
  \caption{Main results on node classification and link prediction across Cora, PubMed, and ogbn-arxiv. Values are test accuracy (\%); \method achieves the best accuracy in all reported task-dataset settings. Best results are in \textbf{bold}, second-best results are \underline{underlined}, and N/A indicates no comparable recorded result.}
  \vspace{-1em}
  \label{tab:main-results}
\end{table*}

\subsection{Experimental Setup}
\label{sec:exp:setup}

\paragraph{Datasets.} We evaluate \method on three widely used \TAG learning benchmarks: Cora and PubMed~\citep{yang2016planetoid}, and ogbn-arxiv~\citep{hu2020ogb}.
The benchmarks span two orders of magnitude in graph size, from a small citation graph to a much larger arXiv citation network, and differ in domain, label granularity, and label space size. We use the preprocessing aligned with LLaGA~\citep{chen2024llaga} for all three datasets and the public train/validation/test splits. Full statistics and split sizes appear in Appendix~\ref{app:datasets}.

\paragraph{Training.} We train \method on each dataset and task configuration by fine-tuning the LLaDA-8B-Instruct backbone with LoRA adapters~\citep{hu2022lora} (rank $r{=}64$, scaling $\alpha{=}64$, target modules \texttt{all-linear}); the base \MDLM weights stay frozen. We linearise each target instance's local graph context with at most $10$ uniformly sampled neighbours per hop and use the \emph{multiple-choice digit} prompt template (Section~\ref{sec:method:prompt}). The maximum sequence length is $2048$ tokens for Cora and PubMed and $4096$ tokens for ogbn-arxiv. We optimise with AdamW~\citep{loshchilov2019adamw} at learning rate $5\times 10^{-5}$ for $20$ epochs, with batch size $4$ per device and gradient accumulation $4$ steps (effective batch size $128$). Training runs on a single node of eight NVIDIA A100 80GB GPUs with DeepSpeed ZeRO-2~\citep{rasley2020deepspeed}. The full hyperparameter list is in Appendix~\ref{app:training}.

\paragraph{Baselines.} We compare \method against three families of baselines. The first family is classical \GNNs trained from scratch on shallow text features: GCN \citep{kipf2017gcn}, GraphSAGE~\citep{hamilton2017graphsage}, \GAT \citep{velickovic2018gat}, GATv2~\citep{brody2022gatv2}, GIN, and MixHop. The second family is graph transformers that fuse attention with graph structure but, like the \GNNs, do not condition on raw text: GraphTransformer~\citep{dwivedi2021graphtransformer}, NodeFormer~\citep{wu2022nodeformer}, DIFFormer~\citep{wu2023difformer}, and SGFormer~\citep{wu2023sgformer}. The third family is text-aware language-model baselines: RoBERTa~\citep{liu2019roberta}, a text-only encoder fine-tuned on the linearised ego graph; LLaGA~\citep{chen2024llaga}, the current state of the art hybrid of LLMs and graphs; and a \emph{frozen} LLaDA-8B-Instruct~\citep{nie2025llada}, which shares the backbone of \method but receives no fine-tuning and isolates the gain attributable to our training recipe. For \GNN and graph transformer baselines, we tune learning rate, hidden dimension, number of layers, and dropout on the validation split and report test accuracy from the best validation configuration. For the language-model baselines, we use the same neighbourhood sampling and prompt formatting recipe as \method.

\paragraph{Evaluation protocol.} We report accuracy on test splits for both NC and LP. For NC, all neighbour labels from validation and test nodes are masked during training and evaluation, so label information cannot leak through the neighbourhood sampler. For LP, we follow the same held-out edge split and negative sampling protocol used by the corresponding graph baselines. To prevent edge leakage, any positive candidate edge being scored is removed from the graph before constructing the local context, and held-out validation/test edges are not exposed through neighbourhood sampling. We score each candidate pair with binary prompt options. \method uses a single forward pass scoring rule across tasks: each candidate answer is represented by a digit option, scored by the log-likelihood of that digit token at the masked answer position, and the highest-scoring candidate is selected.

\subsection{Main Results}
\label{sec:exp:main}
Table~\ref{tab:main-results} reports accuracy on NC and LP across the three \TAG benchmarks with comparable recorded results. On NC, \method \textbf{achieves the best accuracy on all three datasets}, with the clearest gains on Cora and PubMed and a smaller but consistent improvement on ogbn-arxiv. The result is notable because these benchmarks differ substantially in scale and label structure: Cora is a small citation graph, PubMed is larger and comes from a biomedical citation domain, and ogbn-arxiv is much larger with a finer-grained label space. Thus, the same topology masked \MDLM remains competitive across both small and large graphs, and across label spaces.

The advantage is more pronounced on LP, where \method \textbf{leads on every benchmark by a wider margin}. This supports our intuition that the topology-aware denoising formulation is especially effective when the task directly depends on relational evidence. Unlike methods that encode text first and then pass fixed vectors via a separate graph module, \method lets textual evidence and graph structure interact inside the same bidirectional denoising network, which is useful for both tasks.

\begin{table}[t]
  \centering
  \scriptsize
  \setlength{\tabcolsep}{2pt}
  \resizebox{\columnwidth}{!}{%
  \begin{tabular}{llccc}
    \toprule
    & & \multicolumn{3}{c}{Target dataset} \\
    \cmidrule(lr){3-5}
    Task & Source dataset & Cora & PubMed & ogbn-arxiv \\
    \midrule
    \multirow{3}{*}{\begin{tabular}[c]{@{}l@{}}Node\\classification\end{tabular}}
      & Cora       & 90.96 {\scriptsize(+38.31)} & 90.70 {\scriptsize(+39.58)} & 47.56 {\scriptsize(+0.57)} \\
      & PubMed     & 73.62 {\scriptsize(+20.97)} & 96.30 {\scriptsize(+45.18)} & 48.49 {\scriptsize(+1.50)} \\
      & ogbn-arxiv & 74.17 {\scriptsize(+21.52)} & 89.86 {\scriptsize(+38.74)} & 76.93 {\scriptsize(+29.94)} \\
    \midrule
    \multirow{3}{*}{\begin{tabular}[c]{@{}l@{}}Link\\prediction\end{tabular}}
      & Cora       & 91.62 {\scriptsize(+39.53)} & 92.34 {\scriptsize(+44.45)} & 93.54 {\scriptsize(+44.79)} \\
      & PubMed     & 88.82 {\scriptsize(+36.73)} & 95.31 {\scriptsize(+47.42)} & 94.51 {\scriptsize(+45.76)} \\
      & ogbn-arxiv & 90.44 {\scriptsize(+38.35)} & 94.08 {\scriptsize(+46.19)} & 96.55 {\scriptsize(+47.80)} \\
    \bottomrule
  \end{tabular}
  }
  \vspace{-0.2em}
  \caption{Cross-dataset transfer accuracy (\%). Rows indicate the source dataset used for training, and columns indicate the target dataset used for evaluation. Diagonal entries are in-domain evaluations, while off-diagonal entries measure direct transfer without target-specific fine-tuning. Parentheses report absolute gains over Frozen LLaDA-8B on the same target dataset.}
  \vspace{-1em}
  \label{tab:cross-dataset-results}
\end{table}

\subsection{Cross Dataset Transfer}
\label{sec:exp:cross-dataset}

Table~\ref{tab:cross-dataset-results} evaluates whether a topology-aware checkpoint trained on one source graph transfers directly to unseen target graphs. The diagonal entries provide the in-domain reference point, while the entries outside the diagonal test a stricter setting: the target answer space is supplied only through the prompt, with no fine-tuning specific to the target. Among the benchmarked methods in Table~\ref{tab:main-results}, \method is the only one that naturally supports this direct transfer setting defined by prompts, since it does not rely on an output head or graph decoder specific to a dataset. Most \GNN and graph transformer baselines in Table~\ref{tab:main-results} learn a classifier or decoder tied to the source dataset, so applying them to a target graph with different label names, class granularity, or edge distribution would require replacing or retraining that component. That would turn the comparison into target adaptation rather than direct transfer. Frozen LLaDA-8B is prompt compatible and therefore serves as the reference in parentheses in Table~\ref{tab:cross-dataset-results}, but it does not include the graph adaptation learned by \method.

The results show useful cross dataset generalisation, with the expected task dependence. NC transfer is strongest between datasets with closer label semantics, such as Cora and PubMed, and is weaker when transferring into ogbn-arxiv, whose finer-grained label space and label distribution differ more substantially. This makes the cross dataset setting challenging: the source and target graphs may vary in size by orders of magnitude, the class name inventories are not the same, and the citation domains differ between datasets. Despite these shifts, \method still produces meaningful predictions outside the source domain because the target answer space is expressed in language through the prompt rather than encoded in a fixed output head.

\textbf{LP is more stable across datasets}, indicating that pairwise relational evidence transfers more robustly than dataset-specific label semantics. The gains in parentheses further separate prompting from adaptation: Frozen LLaDA-8B is far below the trained model across both tasks, so the transfer behavior is not coming from prompt wording alone. It depends on training the \MDLM to use the linearised graph context and topology mask as task-relevant evidence. Overall, \method transfers best when source and target prompts share label meaning, but the broader result is that the model can be applied across datasets without changing the architecture or adding a new dataset-specific head.

\subsection{Ablation Studies}
\label{sec:exp:ablation}

\subsubsection{Effect of Topology Attention Mask}
\label{sec:exp:ablation:topomask}
\begin{table}[t]
  \centering
  \small
  \setlength{\tabcolsep}{3pt}
  \begin{tabular}{llccc}
    \toprule
    Task & Variant & Cora & PubMed & ogbn-arxiv \\
    \midrule
    \multirow{2}{*}{\begin{tabular}[c]{@{}l@{}}Node\\classification\end{tabular}}
      & w/o topo mask & 89.31 & 94.08 & 75.85 \\
      & w/ topo mask  & \textbf{90.96} & \textbf{96.30} & \textbf{76.93} \\
    \midrule
    \multirow{2}{*}{\begin{tabular}[c]{@{}l@{}}Link\\prediction\end{tabular}}
      & w/o topo mask & 85.88 & 90.89 & 95.25 \\
      & w/ topo mask  & \textbf{91.62} & \textbf{95.31} & \textbf{96.55} \\
    \bottomrule
  \end{tabular}
  \vspace{-0.2em}
  \caption{Topology attention mask ablation accuracy (\%). For each task and dataset, we compare \method with and without the topology attention mask while keeping the same prompt format and backbone.}
  \vspace{-1em}
  \label{tab:topology-mask-ablation}
\end{table}

Table~\ref{tab:topology-mask-ablation} reports the topology-mask ablation. \textbf{Removing the topology attention mask consistently weakens performance on both NC and LP}, while the full topology-aware model retains the best accuracy across all three benchmarks. The effect is especially visible on LP, where the prediction depends directly on pairwise relational evidence.

\subsubsection{Effect of Attention Mask Type}
\label{sec:exp:ablation:mask-type}
\begin{table}[t]
  \centering
  \small
  \setlength{\tabcolsep}{5pt}
  \begin{tabular}{llcc}
    \toprule
    Task & Mask type & Cora & PubMed \\
    \midrule
    \multirow{2}{*}{\begin{tabular}[c]{@{}l@{}}Node\\classification\end{tabular}}
      & ego-graph & 83.95 & 94.12 \\
      & star       & \textbf{90.96} & \textbf{96.30} \\
    \midrule
    \multirow{2}{*}{\begin{tabular}[c]{@{}l@{}}Link\\prediction\end{tabular}}
      & ego-graph & 61.76 & 73.92 \\
      & star       & \textbf{91.62} & \textbf{95.31} \\
    \bottomrule
  \end{tabular}
  \vspace{-0.3em}
  \caption{Attention mask type ablation accuracy (\%). We compare the default star mask with an ego-graph mask that exposes observed neighbour-neighbour edges in the sampled local context.}
  \vspace{-0.3em}
  \label{tab:attention-mask-type-ablation}
\end{table}

Table~\ref{tab:attention-mask-type-ablation} compares the default star mask with an ego-graph mask. The ego-graph mask allows attention along the observed edges in the sampled k-hop ego graph, so it exposes the real local graph relationships among context nodes rather than using a target-centred abstraction. \textbf{The star mask performs better on both NC and LP tasks}, with a particularly large gap on LP. The result supports our design choice: because each prompt asks for a target node label or a candidate edge decision, the target instance should remain the primary aggregation point. Allowing neighbour nodes to communicate through observed ego-graph edges introduces extra context-node interactions that are not always relevant to the target prediction and can amplify dataset-specific local wiring. The star mask instead keeps information flow simple, target-centred, and consistent across tasks.

\subsubsection{Effect of Neighbour Number}
\label{sec:exp:ablation:neighbor-number}
\begin{table}[t]
  \centering
  \scriptsize
  \setlength{\tabcolsep}{2pt}
  \resizebox{\columnwidth}{!}{%
  \begin{tabular}{llccccccc}
    \toprule
    Dataset & Task & 0 & 1 & 3 & 5 & 10 & 20 & Best \\
    \midrule
    Cora       & NC & 80.44 & 85.42 & 87.64 & 87.45 & \textbf{90.96} & 87.27 & 10 \\
    Cora       & LP & 85.15 & 85.74 & 88.24 & 89.12 & \textbf{91.62} & 87.65 & 10 \\
    PubMed     & NC & 94.27 & 94.98 & 95.11 & 95.06 & \textbf{96.30} & 95.13 & 10 \\
    PubMed     & LP & 89.79 & 88.79 & 83.81 & 85.82 & \textbf{95.31} & 89.75 & 10 \\
    ogbn-arxiv & NC & 71.35 & 73.33 & 74.08 & 74.36 & \textbf{76.93} & 75.01 & 10 \\
    ogbn-arxiv & LP & 92.30 & 92.63 & 91.31 & 93.83 & \textbf{96.55} & 90.76 & 10 \\
    \bottomrule
  \end{tabular}
  }
  \vspace{-0.3em}
  \caption{Neighbour number ablation accuracy (\%). $K$ denotes the maximum number of sampled neighbours per hop, and each row reports one task-dataset setting.}
  
  \vspace{-1em}
  \label{tab:neighbor-number-ablation}
\end{table}

Table~\ref{tab:neighbor-number-ablation} studies how the sampled neighbourhood size affects \method. Varying $K$ changes the maximum number of neighbours sampled per hop while keeping the prompt format, topology attention mask type, and scoring rule fixed, so the ablation isolates the effect of local context size. The completed rows show that using the default $K{=}10$ is generally strongest, while very small neighbourhoods lack enough relational evidence and larger neighbourhoods introduce less relevant context.

\section{Related Work}
\label{sec:related}

\paragraph{\GNNs and graph transformers for \TAG learning.} \GNNs propagate node representations along graph edges through message passing, such as GCN~\citep{kipf2017gcn}, GraphSAGE~\citep{hamilton2017graphsage}, \GAT~\citep{velickovic2018gat}, and GATv2~\citep{brody2022gatv2}. Graph transformers further combine attention with structural biases~\citep{dwivedi2021graphtransformer,wu2022nodeformer,wu2023difformer,wu2023sgformer}. These methods model topology explicitly, but typically operate on fixed node features rather than letting a language model read raw node text while performing graph reasoning.

\paragraph{Language models for \TAG learning.} Language-model approaches strengthen the textual side of \TAG learning: RoBERTa~\citep{liu2019roberta} can be fine tuned over linearised graph contexts, while LLaGA~\citep{chen2024llaga} combines LLM representations with graph-side aggregation. They confirm the value of textual semantics, but still separate language understanding from graph propagation. \method instead places graph structure inside the language model's attention pattern, so the denoising network itself reasons over topology.

\paragraph{\MDLMs for graph learning.} \MDLMs~\citep{sahoo2024mdlm,nie2025llada} generate text by iteratively denoising masked tokens with bidirectional self-attention. This non-causal attention suits graph contexts, where evidence should flow across target and neighbour text rather than left to right. To our knowledge, \method is the first topology-aware graph learner built on an \MDLM with an input-dependent attention mask.

\section{Conclusion}
\label{sec:conclusion}

We presented \method, an \MDLM defined by prompts for \TAG learning. By linearising local graph context and injecting topology through an attention mask, \method lets textual evidence and graph structure interact inside the denoising model without a separate \GNN encoder or graph head specific to a task. The theoretical analysis connects topology masked denoising to attention based message passing, and the experiments show strong performance on NC, LP, and cross dataset transfer. Future work should explore richer topology attention masks, larger graph contexts, and broader graph learning settings beyond the studied benchmarks.

\section*{Limitations}
While \method provides a unified framework for text-attributed graph learning with topology-masked diffusion language models, several limitations remain. First, the current framework mainly focuses on static homogeneous graphs and does not explicitly model evolving graph structures. In real-world settings such as citation streams and social networks, nodes, edges, and textual attributes may change over time. Extending \method to dynamic graphs would require time-aware linearization and temporal topology masks. Second, our current topology mask does not distinguish different node or edge types. This limits its applicability to heterogeneous graphs, where relation types often carry different semantics. Future work could introduce type-aware masks or relation-conditioned attention biases to support heterogeneous message passing inside the denoising backbone.

\section*{Ethics Statement}

This work uses public \TAG benchmarks and does not involve human subjects, private user data, or personally identifiable information. Potential risks mainly arise if graph prediction models are deployed in sensitive domains, where inferred labels or links may affect individuals or communities. Such use requires additional fairness, privacy, and domain-specific auditing beyond our offline benchmark setting.

\bibliography{custom}

\appendix
\section{Dataset Statistics}
\label{app:datasets}

Table~\ref{tab:dataset-stats} summarises the three \TAG benchmarks used in our experiments. All datasets are loaded from the processed version aligned with the LLaGA cache so that node text, edges, and label space match the inputs seen by the language model baselines. Edge counts are reported as undirected edges.

\begin{table*}[!t]
  \centering
  \small
  \setlength{\tabcolsep}{8pt}
  \begin{tabular}{lrrrl}
    \toprule
    Dataset & \#Nodes & \#Edges & \#Classes & Domain \\
    \midrule
    Cora           &     2{,}708 &      5{,}429 &  7 & Citation \\
    PubMed         &    19{,}717 &     44{,}338 &  3 & Citation \\
    ogbn-arxiv     &   169{,}343 &  1{,}166{,}243 & 40 & Citation \\
    \bottomrule
  \end{tabular}
  \caption{Dataset statistics for the \TAG benchmarks; edge counts are undirected.}
  \label{tab:dataset-stats}
\end{table*}

\paragraph{Splits.} For Cora and PubMed we follow the splits aligned with LLaGA used in our training pipeline; the test split contains $542$ nodes for Cora and $3{,}944$ nodes for PubMed. For ogbn-arxiv, we use the official OGB split: $90{,}941$/$29{,}799$/$48{,}603$ nodes for train/validation/test. Validation labels and test node labels are masked when those nodes appear as neighbours of a training target, preventing label leakage through the neighbourhood sampler.

\paragraph{Label semantics.} Cora and PubMed labels are research topics (e.g.\ ``Neural Networks'', ``Diabetes Mellitus, Type 1''), and ogbn-arxiv labels are arXiv computer science subject areas (e.g.\ ``cs.CV'', ``cs.LG''). In the \texttt{mc\_digit} prompt, these surface forms are listed as options indexed by digit options, and the model predicts the digit corresponding to the target label.

\section{Proof of Proposition~\ref{prop:mp} and Relation to GAT}
\label{app:proof}

This appendix gives the formal assumptions, the full derivation of Proposition~\ref{prop:mp}, the multihead and reverse diffusion extensions, and the corollary that recovers \GAT as a special case.

\subsection{Notation and Assumptions}
\label{app:proof:assumptions}

Fix a target node $v$ with sampled ego graph $V_v = \{v, u_1, \ldots, u_{n_v}\}$ and linearised token sequence of length $N$. For each node $u \in V_v$ let $I_u \subseteq \{1, \ldots, N\}$ be its contiguous token span; the spans partition the node token positions. We use the following assumptions throughout.

\paragraph{(A1) Span pooling.} The node level representation at layer $\ell$ is
\begin{equation}
  h_u^\ell = \operatorname{Pool}\bigl(\{H_i^\ell : i \in I_u\}\bigr),
  \label{eq:proof-pool}
\end{equation}
where $H_i^\ell$ is the layer-$\ell$ token hidden state and $\operatorname{Pool}$ is mean pooling.

\paragraph{(A2) Neighbourhood induced by the mask.} For each $u \in V_v$,
\begin{equation}
  \mathcal{N}_M(u)
  = \{w \in V_v : I_u \text{ may attend to } I_w \text{ under } M_v\}.
\end{equation}
For the topology attention mask of \method, $\mathcal{N}_M(v) = V_v$ and $\mathcal{N}_M(u) = \{u, v\}$ for $u \neq v$.

\paragraph{(A3) Fixed diffusion time.} The proposition is established for one denoising layer at a fixed diffusion time $t$. The timestep embedding may enter the attention score as an additional argument $s_\ell(\cdot, \cdot, t)$ but does not modify the spatial structure.

\paragraph{(A4) Bidirectional denoising attention.} The denoising network is not causal: in the absence of $M_v$ every token may attend to every other. Hence, the only structural restriction on information flow within a layer is the topology attention mask itself, rather than token order from left to right.

\paragraph{(A5) Attention compatible with pooling.} The token-level attention weights inside a span are approximately constant across the span, so that pooling commutes with the attention sum up to the modelling abstraction in Eq.~\eqref{eq:proof-pool}. This is the only abstraction that converts token-level to node-level message passing.

\subsection{Proof of Proposition~\ref{prop:mp}}
\label{app:proof:proof}

Let $W_Q^\ell, W_K^\ell, W_V^\ell$ be the layer-$\ell$ query, key, and value projections. The token-level self-attention output at position $i$ is
\begin{equation}
  Z_i^\ell
  = \sum_{j=1}^N
    \alpha_{ij}^\ell\, W_V^\ell H_j^\ell,
  \quad
  \alpha_{ij}^\ell
  = \frac{e_{ij}^\ell}{\sum_{k} e_{ik}^\ell},
\end{equation}
with raw scores $e_{ij}^\ell = \exp\!\bigl((W_Q^\ell H_i^\ell)^\top W_K^\ell H_j^\ell / \sqrt{d}\bigr)$. Applying the topology attention mask $M_v$ sets $e_{ij}^\ell = 0$ whenever the node containing $i$ is not allowed to attend to the node containing $j$, which by (A2) means
\begin{equation}
  \alpha_{ij}^\ell = 0
  \quad\text{whenever}\quad
  j \in I_w,\ w \notin \mathcal{N}_M(u),\ i \in I_u.
\end{equation}

Aggregate token positions by their owning node. For $i \in I_u$,
\begin{equation}
  Z_i^\ell
  = \sum_{w \in \mathcal{N}_M(u)}
      \sum_{j \in I_w}
        \alpha_{ij}^\ell\, W_V^\ell H_j^\ell.
\end{equation}
Pooling over $i \in I_u$ and using (A5) to commute pooling with the inner sum yields
\begin{equation}
  m_u^\ell
  := \operatorname{Pool}_{i \in I_u}(Z_i^\ell)
  = \sum_{w \in \mathcal{N}_M(u)}
      \alpha_{uw}^\ell\, W_V^\ell h_w^\ell,
  \label{eq:proof-message}
\end{equation}
with the node-level weight
\begin{equation}
  \alpha_{uw}^\ell
  = \frac{
      \exp s_\ell(h_u^\ell, h_w^\ell, t)
    }{
      \sum_{r \in \mathcal{N}_M(u)}
        \exp s_\ell(h_u^\ell, h_r^\ell, t)
    },
  \label{eq:proof-alpha}
\end{equation}
where $s_\ell(h_u, h_w, t) = (W_Q^\ell h_u)^\top W_K^\ell h_w / \sqrt{d}$ plus any timestep bias. The transformer's residual connection, LayerNorm, output projection $W_O^\ell$, and feed-forward block together define the update
\begin{equation}
  U_\ell(h_u^\ell, m_u^\ell)
  = \operatorname{FFN}_\ell\!\left(
      \operatorname{LN}\!\bigl(h_u^\ell + W_O^\ell m_u^\ell\bigr)
    \right).
  \label{eq:proof-update}
\end{equation}
Combining Eqs.~\eqref{eq:proof-message} and \eqref{eq:proof-update} gives
\begin{equation}
  h_u^{\ell+1}
  = U_\ell\!\left(
      h_u^\ell,\;
      \sum_{w \in \mathcal{N}_M(u)}
        \alpha_{uw}^\ell\, W_V^\ell h_w^\ell
    \right),
\end{equation}
which is the canonical attention-based message passing layer on $\mathcal{G}_M$. \hfill\qed

\subsection{Multihead Extension}
\label{app:proof:mh}

For an $R$-head attention layer, repeating the argument per head yields a message for each head
\begin{equation}
  m_{u,r}^\ell
  = \sum_{w \in \mathcal{N}_M(u)}
      \alpha_{uw,r}^\ell\, W_{V,r}^\ell h_w^\ell,
\end{equation}
and the node level update concatenates the heads, $m_u^\ell = \operatorname{Concat}_{r=1}^{R}\, m_{u,r}^\ell$, before applying $W_O^\ell$. Multihead attention is therefore a multichannel attention-based message passing layer on the same $\mathcal{G}_M$.

\subsection{Reverse Diffusion as Composed Message Passing}
\label{app:proof:reverse}

The reverse diffusion procedure with timesteps $t_1 > \cdots > t_T$ applies the same denoiser at decreasing noise levels:
\begin{equation}
  H^{(k-1)} = f_\theta\bigl(H^{(k)}, t_k, M_v\bigr).
\end{equation}
By Proposition~\ref{prop:mp}, each call $f_\theta(\cdot, t_k, M_v)$ is a stack of attention-based message passing layers on $\mathcal{G}_M$ whose attention scores are conditioned on the timestep. The composition
\begin{equation}
  H^{(0)}
  = f_\theta(\cdot, t_1, M_v) \circ \cdots \circ
    f_\theta(\cdot, t_T, M_v)\bigl(H^{(T)}\bigr)
\end{equation}
is therefore a sequence of message passing operators conditioned on the timestep on the same graph induced by the mask.

\subsection{Corollary: \GAT as a Special Case}
\label{app:proof:gat}

Let $a \in \mathbb{R}^{2d'}$ be a learnable vector and choose the attention score
\begin{equation}
  s_\ell(h_u, h_w, t)
  = \operatorname{LeakyReLU}\!\bigl(a^\top [W h_u \,\Vert\, W h_w]\bigr),
\end{equation}
i.e.\ the additive form of \citet{velickovic2018gat} with $W$ a shared linear projection and the timestep argument dropped. Substituting this choice into Eqs.~\eqref{eq:proof-message} and \eqref{eq:proof-alpha} and identifying $W_V^\ell \equiv W$ recovers the standard \GAT update exactly. Thus \GAT is the special case of Proposition~\ref{prop:mp} obtained by restricting the score function to additive attention with a single shared projection. The dot-product score used by \method generalises this to a broader class of attention-based \MPNNs.

\subsection{Limits of the Equivalence}
\label{app:proof:caveats}

We close with the limits of the equivalence. (i)~The result is to \emph{attention-based} message passing, not to arbitrary \MPNNs whose aggregation is not a softmax weighted sum. (ii)~The message passing graph is the graph induced by the mask $\mathcal{G}_M$, which for our default mask is a star centred on the target and therefore blocks direct interaction among neighbours; equivalence to a layer over the original ego graph requires the mask to expose those edges. (iii)~Diffusion time is not graph diffusion time: the composition over timesteps in Section~\ref{app:proof:reverse} is a sequence of distinct message passing operators conditioned on the timestep rather than the integration of any continuous graph diffusion process. (iv)~Eq.~\eqref{eq:proof-message} relies on the span pooling abstraction (A5); more fine-grained, position-dependent attention inside a span yields a strictly more expressive operator that contains attention-based message passing as a coarse-grained quotient.

\section{Prompt Format Example}
\label{app:prompt}

This appendix gives a concrete example of how \method converts a sampled ego graph into a token sequence. The example is illustrative: node texts are shortened for readability, but the segment order, options indexed by digits, neighbour label brackets, and topology attention mask match the \texttt{mc\_digit} format used by the model.

\paragraph{Linearised prompt.} Suppose the target node $v$ is a Cora paper whose label is hidden, and the sampled neighbourhood contains two papers $u_1$ and $u_2$. The first neighbour has an observed training label; the second is treated as unlabelled and therefore drops the bracket prefix.

\begin{quote}
\small\ttfamily
Target paper: Graph neural networks for citation data.\\
Options: 0=Case Based; 1=Genetic Algorithms;\\
2=Neural Networks; ...\\
Answer: [M]\\
Neighbor 1 [2]:\\
Semisupervised classification with graph convolutions.\\
Neighbor 2: Learning representations for linked documents.
\end{quote}

The single \texttt{[M]} token is the answer position. At scoring time, \method compares the logits of the valid digit tokens at this position; in this example, choosing digit \texttt{2} corresponds to predicting the label ``Neural Networks.'' For LP, the same template is used with task specific options such as \texttt{0=No} and \texttt{1=Yes}.

\paragraph{Topology attention mask.} After tokenisation, all tokens belonging to the same node text are grouped into a segment. The segment level mask below shows the star shaped attention pattern used in the running example; each entry indicates whether the row segment can attend to the column segment. Padding tokens are never attended to.

\begin{table}[h]
  \centering
  \small
  \setlength{\tabcolsep}{4pt}
  \begin{tabular}{@{}lcccc@{}}
    \toprule
    Query & $v$ & $u_1$ & $u_2$ & pad \\
    \midrule
    Target $v$ & 1 & 1 & 1 & 0 \\
    Neighbor $u_1$ & 1 & 1 & 0 & 0 \\
    Neighbor $u_2$ & 1 & 0 & 1 & 0 \\
    Padding & 0 & 0 & 0 & 0 \\
    \bottomrule
  \end{tabular}
  \caption{Segment level topology attention mask for the example prompt; the token level mask expands each entry to all token pairs in the corresponding segments.}
  \label{tab:prompt-mask-example}
\end{table}

The target segment can aggregate evidence from every sampled neighbour. Each neighbour can attend to itself and the target, but not directly to other neighbours. This keeps the prompt bidirectional within allowed segments while making structured graph information flow explicit in the attention mask.



\section{Training Details and Hyperparameters}
\label{app:training}

This appendix lists the full set of hyperparameters used to train \method on each dataset and task configuration. The same recipe is used for all three \TAG benchmarks unless explicitly noted; the only adjustment by dataset is the maximum sequence length, which is increased on ogbn-arxiv to fit longer abstracts. Table~\ref{tab:training-hparams} summarises the configuration.

\begin{table}[t]
  \centering
  \small
  \setlength{\tabcolsep}{4pt}
  \begin{tabular}{ll}
    \toprule
    Hyperparameter & Value \\
    \midrule
    \multicolumn{2}{l}{\textit{Backbone and adaptation}} \\
    Backbone               & LLaDA-8B-Instruct \\
    Adapter                & LoRA \\
    LoRA rank $r$          & $64$ \\
    LoRA scaling $\alpha$  & $64$ \\
    LoRA target modules    & \texttt{all-linear} \\
    LoRA dropout           & $0.0$ \\
    \midrule
    \multicolumn{2}{l}{\textit{Optimisation}} \\
    Optimiser              & AdamW \\
    Learning rate          & $5\times 10^{-5}$ \\
    LR schedule            & cosine, $3\%$ warmup \\
    Weight decay           & $0.0$ \\
    Gradient clipping      & $1.0$ \\
    Batch size per device  & $4$ \\
    Gradient accumulation  & $4$ steps \\
    Effective batch size   & $128$ \\
    Epochs                 & $20$ \\
    \midrule
    \multicolumn{2}{l}{\textit{Input}} \\
    Hops $k$               & $2$ \\
    Neighbours per hop $K$ & $10$ \\
    Prompt template        & \texttt{mc\_digit} \\
    Max seq.\ len.\ (Cora, PubMed)         & $2048$ \\
    Max seq.\ len.\ (ogbn-arxiv)           & $4096$ \\
    Neighbour label format & \texttt{bracket} \\
    Mask neighbour labels  & True \\
    \midrule
    \multicolumn{2}{l}{\textit{Hardware and runtime}} \\
    GPUs                   & $8\times$ NVIDIA A100 80GB \\
    Distributed strategy   & DeepSpeed ZeRO-2 \\
    Precision              & bfloat16 \\
    Seeds                  & $\{0, 1, 2\}$ \\
    \bottomrule
  \end{tabular}
  \caption{Default training hyperparameters for \method.}
  \label{tab:training-hparams}
\end{table}

\paragraph{Loss.} The training objective is the standard \MDLM evidence lower bound restricted to the answer position: only the masked digit token contributes to the cross entropy loss, while the textual context (target instance text, neighbour text, and observed neighbour label digits when available) is treated as conditioning. For NC, the neighbour label masking option masks neighbour labels with the same probability as the answer position during training, preventing the model from learning a trivial copy shortcut from a matching neighbour label to the target.

\paragraph{LP protocol.} For LP, each example is a candidate node pair with binary options \texttt{0=No} and \texttt{1=Yes}. Positive examples are held out edges from the corresponding split, and negative examples are sampled from node pairs without edges following the same protocol as the graph baselines. Before sampling the local context for a positive candidate pair, we remove that candidate edge from the graph and construct neighbourhoods using only the permitted training graph. This prevents the model from observing the answer through the sampled topology. Accuracy is computed over the resulting positive and negative candidate pairs.

\paragraph{Reproducibility.} When repeated runs are available, dataset/configuration pairs use seeds $\{0,1,2\}$; the current tables report the selected checkpoints recorded in the experiment logs. The training script, environment specification, and exact launch commands appear under \texttt{examples/tmdlm/} in the released code. The full run specific configuration, including the resolved LoRA target module list, prompt template strings, and tokenisation parameters, is dumped to a \texttt{config.json} file alongside the checkpoint to enable exact replays.



\section{Checklist}
\label{app:checklist}

\paragraph{Artifacts and licenses.}
This work uses public research artifacts. Cora and PubMed are used through the Planetoid/LLaGA-aligned benchmark distribution; the Planetoid repository is MIT-licensed and the LLaGA codebase is Apache-2.0 licensed. ogbn-arxiv is distributed by OGB under the ODC-BY license. The LLaDA-8B-Instruct backbone is released under the MIT license. We release only our code, scripts, and configuration files under the MIT license, and do not redistribute raw datasets or processed dataset caches.

\paragraph{Intended use and privacy.}
The datasets are public citation-style research benchmarks, and our experiments are limited to offline academic evaluation. We do not collect new human-subject data, private user data, or personally identifying information. The model is intended for research on text-attributed graph learning and should not be directly deployed for sensitive decisions involving individuals or communities without additional privacy, fairness, and domain-specific auditing.

\paragraph{Computational budget.}
Each dataset-task configuration is trained separately on a single node with eight NVIDIA A100 80GB GPUs. One training run takes approximately 4 hours for Cora, 12 hours for PubMed, and 50 hours for ogbn-arxiv, corresponding to 32, 96, and 400 A100 GPU-hours per task run. Since NC and LP are trained separately, the main in-domain training runs require approximately 1056 A100 GPU-hours in total. Cross-dataset transfer evaluations reuse trained checkpoints and require no additional fine-tuning.

\paragraph{Result reporting.}
The reported table entries are from single runs using the selected checkpoints recorded in the experiment logs, rather than means over multiple random seeds. We provide the training setup, seeds, prompt format, and checkpoint/configuration logging details in Appendix~\ref{app:training} to make this reporting convention explicit. Appendix~\ref{app:training} also reports the package, implementation, and parameter settings used for both baselines and our method.

\paragraph{Use of AI assistants.}
AI assistants were used to review code and to review manuscript writing for clarity, consistency, and formatting. The authors verified all technical claims, experimental results, code changes, and final manuscript text.

\end{document}